%% file: kernels_arxiv.tex
\documentclass[11pt,table]{article}
\pdfoutput=1

\usepackage[utf8]{inputenc} 
\usepackage[T1]{fontenc}    
\usepackage[in]{fullpage}
\usepackage{xurl}
\usepackage{booktabs}       
\usepackage{amsfonts}       
\usepackage{nicefrac}       
\usepackage{microtype}      
\usepackage{amsmath,amssymb,amsthm}
\usepackage{mathtools}
\usepackage{mathrsfs}
\usepackage{thmtools}
\usepackage{thm-restate}
\usepackage{comment}
\usepackage{parskip}
\usepackage{siunitx}
\usepackage{authblk}
\usepackage{etoolbox}
\usepackage{enumitem}
\usepackage{graphicx}
\usepackage{subcaption}
\usepackage[textsize=scriptsize,textwidth=2.2cm]{todonotes}
\usepackage{tabularx}
\usepackage{xcolor} 
\usepackage[numbers,sort]{natbib}
\usepackage{hyperref}
\usepackage{longtable}
\usepackage{afterpage}
\usepackage{algorithm}
\usepackage{algpseudocode}
\usepackage{bm}
\usepackage{etoc}

\extrafloats{100}

\newtoggle{showtodos}
\togglefalse{showtodos}

\newtoggle{showappendix}
\toggletrue{showappendix}

\newtoggle{isicml}
\togglefalse{isicml}

\newcommand{\R}{\mathbb{R}}
\newcommand{\E}{\mathbb{E}}

\newcommand{\vct}[1]{\bm{#1}}
\newcommand{\mtx}[1]{\bm{#1}}
\hypersetup{draft}
\algblock{Input}{EndInput}
\algnotext{EndInput}
\algblock{Output}{EndOutput}
\algnotext{EndOutput}
\newcommand{\Desc}[2]{\State \makebox[2em][l]{#1}#2}

\newcommand{\cifarten}{CIFAR-10}
\usepackage[title]{appendix}

\title{Neural Kernels Without Tangents}
\author[1]{Vaishaal Shankar}
\author[1]{Alex Fang}
\author[1]{Wenshuo Guo}
\author[1]{Sara Fridovich-Keil}
\author[1]{Ludwig Schmidt}
\author[2]{Jonathan Ragan-Kelley}
\author[1]{Benjamin Recht}
\affil[1]{University of California, Berkeley}
\affil[2]{MIT CSAIL}

\begin{document}

\date{}

\maketitle

\begin{abstract}
    \input{abstract}

\end{abstract}

\section{Introduction}

\input{introduction}

\section{Related Work}
\input{related_work}

\label{sec:related_work}

\section{Compositional kernels for bags of features}
\label{sec:kernels}
\input{kernel}

\section{Experiments}
\label{sec:experiments}
\input{experiments}

\section{Limitations and Future Work}
\label{sec:conclusion}
\input{conclusion}

\section*{Acknowledgements}
\input{acknowledgements}

\label{submission}

\bibliography{cites}
\bibliographystyle{icml2020}

\clearpage
\begin{appendices}

\section{Nonparametric prediction with kernels}
\label{app:background}
\input{app_background}

\section{LOO Tilting and ZCA Augmentation}
\label{app:loozca}
\input{app_loozca}

\section{Supplementary proof}
\label{app:proof}
\input{app_proof}

\section{Neural Network Parameters}
\label{app:params}
\input{app_params}

\section{Neural Network Architectures}
\label{app:params}
\input{app_archs}

\section{Subsampled {\cifarten} experiments details}
\label{app:subcifar}
\input{app_subcifar}

\section{UCI experiments details}
\label{app:UCI}
\input{app_UCI}

\end{appendices}

\end{document}

%% file: abstract.tex
%

We investigate the connections between neural networks and simple building blocks in kernel space. In particular, using well established feature space tools such as direct sum, averaging, and moment lifting, we present an algebra for creating ``compositional" kernels from bags of features. We show that these operations correspond to many of the building blocks of ``neural tangent kernels" (NTK). Experimentally, we show a correlation in test error between neural network architectures and the associated kernels. We construct a simple neural network architecture using only $3 \times 3$ convolutions, $2 \times 2$ average pooling, ReLU, and optimized with SGD and MSE loss that achieves $96\%$ accuracy on CIFAR10, and whose corresponding compositional kernel achieves $90\%$ accuracy. We also use our constructions to investigate the relative performance of neural networks, NTKs, and compositional kernels in the small dataset regime. In particular, we find that compositional kernels outperform NTKs and neural networks outperform both kernel methods.

%% file: introduction.tex
Recent research has drawn exciting connections between neural networks and kernel methods, providing new insights into training dynamics, generalization, and expressibility~\cite{daniely2016toward, ntk, Du19, Du19b, ghorbani2019linearized, allen2019learning, lee2019wide}. This line of work relates ``infinitely wide'' neural networks to particular kernel spaces, showing that infinite limits of random initializations of neural networks lead to particular kernels on the same input data. Since these initial investigations, some have proposed to use these kernels in prediction problems, finding promising results on many benchmark problems \cite{li2019enhanced, arora2020harnessing}. However, these kernels do not match the performance of neural networks on most tasks of interest, and the kernel constructions themselves are not only hard to compute, but their mathematical formulae are difficult to even write down~\cite{cntkexact}.

In this paper, we aim to understand empirically if there are computationally tractable kernels that approach the expressive power of neural networks, and if there are any practical links between kernel and neural network architectures. We take inspiration from both the recent literature on ``neural tangent kernels'' (NTK) and the classical literature on compositional kernels, such as ANOVA kernels. We describe a set of three operations in feature space that allow us to turn data examples presented as collections of small feature vectors into a single expressive feature-vector representation. We then show how to compute these features directly on kernel matrices, obviating the need for explicit vector representations. We draw connections between these operations, the compositional kernels of \citet{daniely2016toward}, and the Neural Tangent Kernel limits of \citet{ntk}. These connections allow us to relate neural architectures to kernels in a transparent way, with appropriate simple analogues of convolution, pooling, and nonlinear rectification (Sec.~\ref{sec:kernels}).

Our main investigation, however, is not in establishing these connections. Our goal is to test whether the analogies between these operations hold in practice: is there a correlation between neural architecture performance and the performance of the associated kernel? Inspired by simple networks proposed by  David \citet{myrtle}, we construct neural network architectures for computer vision tasks using only $3 \times 3$ convolutions, $2 \times 2$ average pooling, and ReLU nonlinearities. We show that the performance of these neural architectures on {\cifarten} strongly predicts the performance of the associated kernel. The best architecture achieves  $96\%$ accuracy on {\cifarten} when trained with SGD on a mean squared error (MSE) loss. The corresponding compositional kernel achieves $90\%$ accuracy, which is, to our knowledge, the highest accuracy achieved thus far by a kernel machine on {\cifarten}. We emphasize here that we compute an \emph{exact kernel} directly from pixels, and do not rely on random feature approximations often used in past work. 

On {\cifarten}, we observe that compositional kernels provide dramatically better results than Neural Tangent Kernels. We also demonstrate that this trend holds in the ``small data'' regime~\cite{arora2020harnessing}. Here, we find that compositional kernels outperform NTKs and neural networks outperform both kernel methods when properly tuned and trained. On a benchmark of 90 UCI tabular datasets, we find that simple, properly tuned Gaussian kernels perform, on aggregate, slightly better than NTKs. Taken together, our results provide a promising starting point for designing practical, high performance, domain specific kernel functions \footnote{code used to produce our results can be found at: \url{https://github.com/modestyachts/neural_kernels_code}}. We suggest that while some notion of compositionality and hierarchy may be necessary to build kernel predictors that match the performance of neural networks, NTKs themselves may not actually provide particularly useful guides to the practice of kernel methods.

%


%% file: related_work.tex
We build upon many prior efforts to design specialized kernels that model specific types of data. 
In classical work on designing kernels for pattern analysis, \citet{shawe2004kernel} establishes an algebra for constructing kernels on structured data. 
In particular, we recall the construction of the ANOVA kernels, which are defined recursively using a set of base kernels. 
Many ideas from ANOVA kernels transfer naturally to images, with operations that capture the similarities between different patches of data.

More recent work \cite{mairal2014convolutional} proposes a multi-layer ``convolutional kernel network" (CKN) for image classification that iterates convolutional and nonlinear operations in kernel space. 
While the structure of our compositional kernels is similar to CKNs, \citet{mairal2014convolutional} are unable to explore networks deeper than two layers due to the inefficiency of their construction, which involves approximating a kernel feature map using optimization. 
In contrast, we compute our kernel \emph{exactly}, and the complexity of our compositional kernel functions is worst-case linear in depth, enabling us to explore much deeper kernel compositions.

Another line of recent work investigates the connection between kernel methods and infinitely wide neural networks. \citet{ntk} posits that
least squares regression with respect to the neural tangent kernel (NTK) is equivalent to optimizing an infinitely wide neural network with gradient flow. 
Similarly, it has been shown that optimizing just the \emph{last} layer of an infinitely wide neural network is equivalent to a Gaussian process (NNGP) based on the neural network architecture \cite{lee2018deep}. Both of these equivalences extend to convolutional neural networks (CNNs) \cite{li2019enhanced, novak2019bayesian}.  
The compositional kernels we explore can be expressed as NNGPs.

To construct our compositional kernel functions, we rely on key results from \citet{daniely2016toward}, which explicitly studies the duality between neural network architectures
and compositional kernels.

%% file: kernel.tex
\label{ss:compbagfeatures}

A variety of data formats are naturally represented by collections of related vectors. For example, an image can be considered a spatially arranged collection of 3-dimensional vectors. A sentence can be represented as a sequence of word embeddings. Audio can be represented as temporally ordered short-time Fourier transforms. In this section, we propose a generalization of these sorts of data types, and a set of operations that allow us to compress these representations into vectors that can be fed into a downstream prediction task. We then show how these operations can be expressed as kernels and describe how to compute them. None of the operations described here are novel, but they form the basic building blocks that we use to build classifiers to compare to neural net architectures.

A \emph{bag of features} is simply a generalization of a matrix or tensor: whereas a matrix is a list of vectors indexed by the natural numbers, a bag of features is a collection of elements in a Hilbert space $\mathcal{H}$ with a finite, structured index set $\mathcal{B}$. As a canonical example, we can consider an image to be a bag of features where the index set $\mathcal{B}$ is the pixel's row and column location and $\mathcal{H}$ is $\R^3$: at every pixel location, there is a corresponding vector in $\R^3$  encoding the color of that pixel. In this section we will denote a generic bag of features by a bold capital letter, e.g., $\mtx{X}$, and the corresponding feature vectors by adding subscripts, e.g., $\mtx{X}_b$. That is, for each index $b \in \mathcal{B}$, $\vct{X}_b\in \mathcal{H}$. 

If our data is represented by a bag of features, we need to map it into a single Hilbert space to perform linear (or nonlinear) predictions. We describe three simple operations to compress a bag of features into a single feature vector.\\

\paragraph{Concatenation.} Let $\mathcal{S}_1,\ldots, \mathcal{S}_L \subseteq \mathcal{B}$ be \emph{ordered} subsets with the same cardinality, $s$. We write each subset as an ordered set of indices: $\mathcal{S}_j = \{i_{j1},\ldots,i_{js}\}$. Then we can define a new bag of features $c(\mtx{X})$ with index set $\{1,\ldots,L\}$ and Hilbert space $\mathcal{H}^s$ as follows. For each $j=1,\ldots, L$, set
\[
	c(\mtx{X})_j = (\mtx{X}_{i_{j1}},\mtx{X}_{i_{j2}},\ldots,\mtx{X}_{i_{js}})\,.
\]
The simplest concatenation is setting $\mathcal{S}_1=\mathcal{B}$, which corresponds to vectorizing the bag of features. As we will see, more complex concatenations have strong connections to convolutions in neural networks.

\paragraph{Downsampling.} Again let $\mathcal{S}_1,\ldots, \mathcal{S}_L \subseteq \mathcal{B}$ be subsets, but now let them have arbitrary cardinality and order. We can define a new bag of features $p(\mtx{X})$ with index set $\{1,\ldots,L\}$ and Hilbert space $\mathcal{H}$. For each $j=1,\ldots, L$ set
\[
	p(\mtx{X})_j = \frac{1}{|\mathcal{S}_j|} \sum_{i \in \mathcal{S}_j}\mtx{X}_{i}\,.
\]
This is a useful operation for reducing the size of $\mathcal{B}$. Here we use the letter $p$ for the operation as downsampling is commonly called ``pooling'' in machine learning.

\paragraph{Embedding.} Embedding simply means a isomorphism of one Hilbert space to another. Let $\varphi:\mathcal{H} \rightarrow \mathcal{H}'$ be a map. Then we can define a new bag of features $\Phi(\mtx{X})$ with index set $\mathcal{B}$ and Hilbert Space $\mathcal{H}'$ by setting
\[
	\Phi(\mtx{X})_b = \varphi(\mtx{X}_b)\,.
\]
Embedding functions are useful for increasing the expressiveness of a feature space.

\subsection{Kernels on bags of features}
Each operation on a bag of features can be performed directly on the kernel matrix of all feature vectors. Given two bags of features with the same $(\mathcal{B},\mathcal{H})$, we define the kernel function
\[
	k(\mtx{X},a,\mtx{Z},b) = \langle \mtx{X}_a, \mtx{Z}_b \rangle\,.
\]
Note that this implicitly defines a \emph{kernel matrix} between two bags of features: we compute the kernel function for each pair of indices in $\mathcal{B} \times \mathcal{B}$ to form a $|\mathcal{B}|\times |\mathcal{B}|$ matrix. Let us now describe how to implement each of the above operations introduced in Section \ref{ss:compbagfeatures}.\\

\paragraph{Concatenation.} Since
\[
	\langle c(\mtx{X})_j, c(\mtx{Z})_k \rangle = \sum_{\ell =1}^s \langle \mtx{X}_{i_{j\ell}}, \mtx{Z}_{i_{k\ell}} \rangle\,,
\]
we have
\[
	k(c(\mtx{X}), j,c(\mtx{Z}), k) = \sum_{\ell =1}^s  k( \mtx{X}, {i_{j\ell}}, \mtx{Z}, {i_{k\ell}} )\,.
\]

\paragraph{Downsampling.} Similarly, for downsampling, we have
\[
		k(c(\mtx{X}), j,c(\mtx{Z}), k) = \frac{1}{|\mathcal{S}_j||\mathcal{S}_k|}\sum_{i \in \mathcal{S}_j} \sum_{\ell \in \mathcal{S}_k} k( \mtx{X}, i, \mtx{Z}, \ell)\,.
\]

\paragraph{Embedding.} Note that the embedding function $\varphi$ induces a kernel on $\mathcal{H}$. If $\vct{x}$ and $\vct{z}$ are elements of $\mathcal{H}$, define
\[
		k_\varphi(\vct{x},\vct{z}) = \langle \varphi(\vct{x}),\varphi(\vct{z}) \rangle\,.
\]
Then, we don't need to materialize the embedding function to compute the effect of embedding a bag of features. We only need to know $k_\varphi$:
\begin{equation}\label{eq:kernel-embedding-operation}
	k(\Phi(\mtx{X}), j,\Phi(\mtx{Z}), k) = k_\varphi(\mtx{X}_j,\mtx{Z}_k)\,.
\end{equation}
We will restrict our attention to  $\varphi$ where we can  compute $k_\varphi(\vct{x},\vct{z})$ only from $\langle \vct{x},\vct{z}\rangle$, $\|\vct{x}\|$ and $\|\vct{z}\|$. 
This will allow us to iteratively use Equation~\eqref{eq:kernel-embedding-operation} in cascades of these primitive operations.

\subsection{Kernel operations on images}
\label{ss:kernels_images}
In this section, we specialize kernel operations to operations on images.
 As described in Section \ref{ss:compbagfeatures}, images are bags of three dimensional vectors indexed by two spatial coordinates. 
Assuming that our images have $D_1\times D_2$ pixels, we create a sequence of kernels by composing the three operations described above.

\paragraph{Input kernel.} The input kernel function $k_{0}$ relates all pixel vectors between all pairs of images in our dataset. 
Computationally, given $N$ images, we can use an image tensor $\bm{T}$ of shape $N \times D_1 \times D_2 \times 3$ to represent the whole dataset of images, and map this into a kernel tensor $\bm{K}_{out}$ of shape $N \times D_1 \times D_2 \times N \times D_1 \times D_2$. The elements of
$\bm{K}_{out} = k_{0}(\bm{T})$ can be written as:
\begin{align*}
    K_{out}[i,j,k,\ell,m,n] = \langle T[i,j,k], T[\ell,m,n]\rangle \,.
\end{align*}
All subsequent operations operate on 6-dimensional tensors with the same indexing scheme.

\paragraph{Convolution.}
The convolution operation $c_w$ maps an input tensor $\bm{K}_{in}$ 
to an output tensor $\bm{K}_{out}$ of the same shape: $N \times D_1 \times D_2 \times N \times D_1 \times D_2$. 
$w$ is an integer denoting the size of the convolution (e.g. $w = 1$ denotes a $3 \times 3$ convolution).

The elements of $\bm{K}_{out} = c_w(\bm{K}_{in})$ can be written as:
\begin{align*}
\begin{split}
    &\quad K_{out}[i,j,k,\ell,m,n]=  \\
    &\sum_{dx=-w}^{w} \sum_{dy=-w}^w K_{in}[i,j+dx,k+dy,\ell,m+dx,n+dy]
\end{split}
\end{align*}
For out-of-bound location indexes, we simply zero pad the $\bm{K}_{in}$ so all out-of-bound accesses return zero.

\paragraph{Average pooling.}
The average pooling operation $p_w$ downsamples the spatial dimension, mapping an input tensor $\bm{K}_{in}$ of shape $N \times D_1 \times D_2 \times N \times D_1 \times D_2$
to an output tensor $\bm{K}_{out}$ of shape $N \times (D_1/w) \times (D_2/w) \times N \times (D_1/w) \times (D_2/w)$. 
We assume $D_1$ and $D_2$ are divisible by $w$.

The elements of $\bm{K}_{out} = p_w(\bm{K}_{in})$ can be written as:
\begin{align*}
\begin{split}
    & K_{out}[i,j,k,\ell,m,n] =\frac{1}{w^4} \sum_{a=1}^{w}\sum_{b=1}^{w}\sum_{c=1}^{w}\sum_{d=1}^{w} \\
    & \bigg(K_{in} [i,wj + a, wk+b,\ell,wm+c,wn+d]\bigg)
\end{split}
\end{align*}

\paragraph{Embedding.}
The nonlinearity layers add crucial nonlinearity to the kernel function, without which the entire map would be linear 
and much of the benefit of using a kernel method would be lost. 
We first consider the kernel counterpart of the ReLU activation.

The ReLU embedding, $k_{relu}$, is shape preserving, mapping an input tensor $\bm{K}_{in}$ of shape $N \times D_1 \times D_2 \times N \times D_1 \times D_2$
to an output tensor $\bm{K}_{out}$ of shape $N \times D_1 \times D_2 \times N \times D_1 \times D_2$. To ease the notation, we define two auxiliary tensors: $\bm{A}$ with shape  $N \times D_1 \times D_2$ and $\bm{B}$ with shape $N \times D_1 \times D_2 \times N \times D_1 \times D_2$, where the elements of each are:
\begin{align*}
A[i, j, k] & = \sqrt{K_{in}[i,j,k,i,j,k]}\\
B [i, j, k, \ell, m, n] &= \arccos \left(\frac{K_{in}[i,j,k,\ell,m,n]}{A[i,j,k] A[\ell, m, n]}\right)
\end{align*}
The elements of $\bm{K}_{out} = k_{relu}(\bm{K}_{in})$ can be written as:
\begin{align*}
\begin{split}
     &K_{out}[i,j,k,\ell,m,n] \\
     =\;&\frac{1}{\pi}\bigg(A[i, j, k] A[\ell,m, n] \sin (B[i, j, k, \ell, m, n])+ \\
       &(\pi - B[i, j, k, \ell, m, n]) \cos (B[i, j, k, \ell, m, n])\bigg)
\end{split}
\end{align*}
The relationship between the ReLU operator and the ReLU kernel is covered in Subsection \ref{ss:nn_kernel}.

In addition to the ReLU kernel, we also work with a normalized Gaussian kernel. The elements of $\bm{K}_{out} = k_{gauss}(\bm{K}_{in})$ can be written as:
\begin{align*}
\begin{split}
     &K_{out}[i,j,k,\ell,m,n] \\
     =\;&A[i, j, k] A[\ell, m, n] \exp( B[i, j, k, \ell, m, n]  -1)
\end{split}
\end{align*}

\begin{figure}
\centering
   \includegraphics[width=0.48\textwidth]{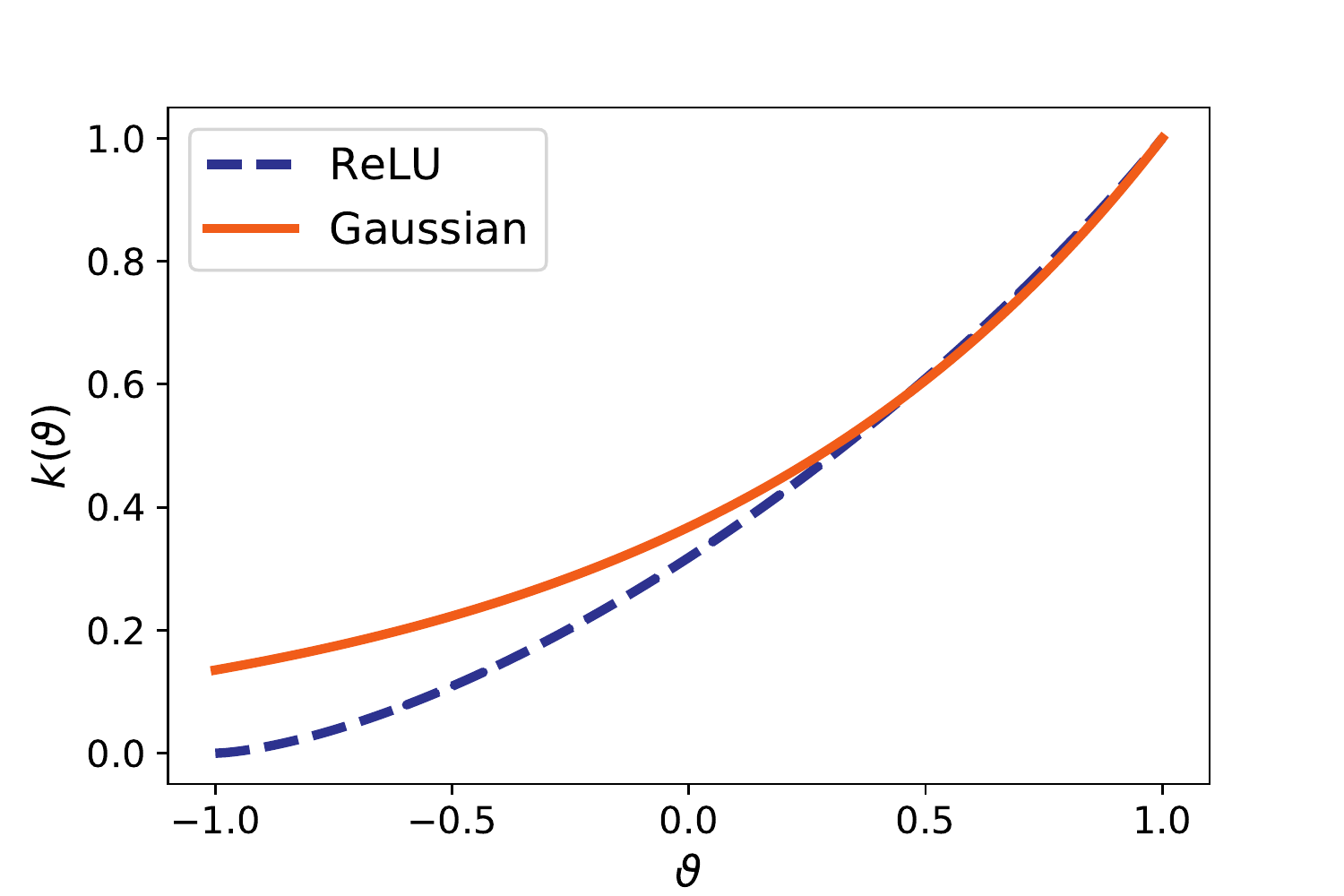}
   \caption{Comparison of the ReLU (arccosine) and Gaussian kernels ($\gamma = 1$), as a function of the angle $\vartheta$ between two examples.} \label{fig:nonlinearities}
\end{figure}

The normalized Gaussian kernel has a similar output response to the ReLU kernel (shown in Figure \ref{fig:nonlinearities}). Experimentally, we find the Gaussian kernel to be marginally faster and more numerically stable. 

\subsection{Relating compositional kernels to neural network architectures}
\label{ss:nn_kernel}
Each of these compositional kernel operations is closely related to neural net architectures, with close ties to the literature on random features \cite{rahimi2008random}. Consider two tensors: $\bm{U}$ of shape $N \times D_1\times D_2 \times D_3$ and $\bm{W}$ of shape $(2w+1) \times (2w+1) \times D_3 \times D_4$. $\bm{U}$ is the input, which can be N images, $w$ is an integer denoting the size of the convolution (e.g. $w = 1$ denotes a $3 \times 3$ convolution), and $\bm{W}$ is a tensor contains the ``weights'' of a convolution. Consider a simple convolutional layer followed by a ReLU layer in a neural network:
\[
	\Psi(\bm{U})= \operatorname{relu} ( \bm{W} * \bm{U})
\]
where ``$*$" denotes the convolution operation and $\operatorname{relu}$ denotes elementwise ReLU nonlinearity. 

A convolution operation can be rewritten as a matrix multiplication with a reshaping of input tensors. We first flatten the weights tensor $\bm{W}$ to a matrix $\bm{W'}$  of $D_{4}$ rows and  $D_{3}(2w+1)^{2}$ columns. For the input tensor $\bm{U}$, given the convolution size $(2w+1) \times (2w+1)$, we consider the ``patch" of each entry $U[n, d_1, d_2, c]$ , which includes the $(2w+1) \times (2w+1)$ entries $U[n, i, j, c]$, where $i \in [d_1-w, d_1 + w], ~j \in [d_2-w, d_2 + w]$. Therefore, we can flatten the input tensor $\bm{U}$ to a matrix $\bm{U'}$ of size $D_{3}(2w+1)^{2} \times D_{1}D_{2}N$ by padding all out-of-bounds entries in the patches to zero. 

The ReLU operation is shape preserving, applying the ReLU nonlinearity $\varphi(x)$ elementwise to the tensor. Thus we can rewrite the above convolution and ReLU operations into
\begin{align*}
    \Psi(\bm{U}) = \operatorname{relu}(\bm{W'}\bm{U'}) = \operatorname{relu}(\bm{W} * \bm{U}) 
\end{align*}

Therefore, a simple convolution layer and a ReLU layer give us an output tensor $\Psi(\bm{U})$ of shape $N \times D_{1} \times D_{2} \times D_{4}$.




With the help of random features, we are able to relate the above neural network architecture to kernel operations. Suppose the entries of $\bm{W}$ are appropriately scaled random Gaussian variables. We can evaluate the following expectation according to the calculation in \citet{daniely2016toward}, thereby relating our kernel construction to inner products between the outputs of \emph{random} neural networks:
\begin{align}
        \begin{split}
        &\mathbb{E}\bigg[ \sum_{c=1}^{D_4} \Psi(\bm{U})[i,j,k,c] \Psi(\bm{U})[\ell,m,n,c]\bigg] =\\ 
        &k_{relu}\Big(c_w\big(k_{0}(\bm{U})\big)\Big)[i,j,k,\ell,m,n]
        \end{split} \label{eq:daniely}
\end{align}
where $k_{0}$ is the input kernel defined in Subsection \ref{ss:kernels_images}. 
We include the proof for the above equality in the appendix.


Similar calculations can be made for the pooling operation, and for any choice of nonlinearity for which the above expectation can be computed. Moreover, since in Eq (\ref{eq:daniely}), the term inside the expectation only depends on inner products, this relation can be generalized to arbitrary depths. 

\subsection{Implementation}
Now we actualize the above formulations into a procedure to generate a kernel matrix from the input data.
Let $\mathcal{A}$ be a set of valid neural network operations.
A given network architecture $\mathcal{N}$ is represented as an ordered list of operations from $\mathcal{A}$.
Let $\mathcal{K}$ denote a mapping from elements of $\mathcal{A}$ to their corresponding operators as defined in Subsection \ref{ss:kernels_images}.

Algorithm \ref{algo:kernel} defines the procedure for constructing a compositional kernel from a given architecture 
$\mathcal{N}$ and an input tensor $\bm{X}$ of $N$ RGB images of shape $N \times D \times D \times 3$. We note that the output kernel is only a $N \times N$ matrix if there exist exactly $\log D$
pooling layers. We emphasize that this procedure is a deterministic function of the input images and network architecture.

Due to memory limitations, in practice we compute the compositional kernel in batches on a GPU. Implementation details are given in Section \ref{sec:experiments}.

\begin{algorithm}
    \begin{algorithmic}
      \Input
      \Desc{$\mathcal{N}$}{Input architecture of $m$ layers from $\mathcal{A}$}
      \Desc{$\mathcal{K}$}{Map from $\mathcal{A}$ to layerwise operators}
      \Desc{$\bm{X}$}{Tensor of input images, shape $(N \times D \times D \times 3)$ }
      \EndInput
      \Output
      \Desc{$\bm{K}_{m}$}{Compositional kernel matrix, shape $(N \times N)$}
      \EndOutput
      \State $\bm{K}_{0} = k_{0}(\bm{X})$
      \For{$i =1$ \textbf{to} $m$}
        \State $k_{i} \gets \mathcal{K}(\mathcal{N}_{i})$
        \State $\bm{K}_{i} \gets k_{i}(\bm{K}_{i - 1})$
      \EndFor 
    \end{algorithmic}
    \caption{Compositional Kernel}\label{algo:kernel}
\end{algorithm}

%% file: experiments.tex
\begin{figure*}[h]
    \centering
    \includegraphics[width=\textwidth]{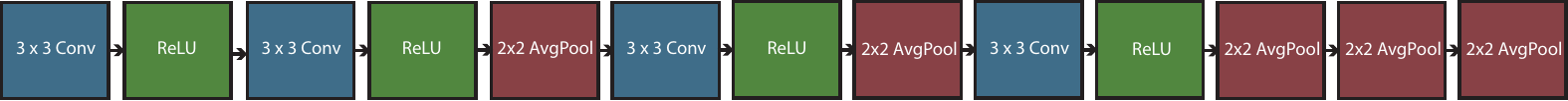}
    \caption{\label{fig:myrtle5} A 5 layer network from the ``Myrtle'' family (Myrtle5). }
\end{figure*}

In this section, we first provide an overview of the architectures used in our experiments. We then present comparison results between neural networks, NTKs, and compositional kernels on a variety of datasets, including MNIST, {\cifarten} (\citet{krizhevsky2009learning}), {\cifarten}.1 (\citet{recht2019imagenet}), {\cifarten}0 (\citet{krizhevsky2009learning}) and 90 UCI datasets (\citet{fernandez2014we}).

\subsection{Architectures}
We design our deep convolutional kernel based on the non-residual convolutional ``Myrtle" networks introduced in \citet{myrtle}. 
We choose this particular network because of its rare combination of simplicity and high performance.
Many components commonly used in neural networks, including residual connections, are intended to ease training but have little or unclear effect in terms of the function of the trained network. 
It is unclear how to model these neural network components in the corresponding kernels, but equally unclear what benefit this might offer.
We further simplify the architecture by removing batch normalization and swapping out max pooling with average pooling, for similar reasons.
The remaining components are exclusively $3 \times 3$ convolutions, $2 \times 2$ average pools, and ReLUs.
More generally, we refer to all architectures that can be represented as a list
of operations from the set \{\verb|conv3|, \verb|pool2|, \verb|relu|\} as the ``Myrtle'' family.

We work with 3 networks from this family: Myrtle5, Myrtle7 and Myrtle10, denoting the depth of each network. An example of the Myrtle5 architecture is shown in Figure \ref{fig:myrtle5}.
The deeper variants have more convolution and ReLU layers;
we refer the reader to the appendix for an illustration of the exact architectures. Next we show convolutional neural networks from this family can indeed achieve high accuracy on {\cifarten}, as can their kernel counterparts.

\subsection{Experimental setup.}
We implemented all the convolutional kernels in the tensor comprehensions framework \cite{vasilache2018tensor} and executed 
them on V100 GPUs using Amazon Web Services (AWS) P3.16xlarge instances. For image classification tasks (MNIST, {\cifarten}, {\cifarten}.1, and {\cifarten}0), we used compositional kernels based on the Myrtle family described above. For tabular datasets (90 UCI datasets), we used simpler Gaussian kernels. All experiments on {\cifarten}, {\cifarten}.1 and {\cifarten}0 used ZCA whitening as a preprocessing step, except for the comparison experiments explicitly studying preprocessing. We apply ``flip" data augmentation to our kernel method by flipping every example in the training set across the vertical axis and constructing a kernel matrix on the concatenation of the flipped and standard datasets. 

For all image classification experiments (MNIST, \cifarten, \cifarten.1, and \cifarten0) we perform kernel ridge regression with respect to one-hot labels, and solve the optimization problem exactly using a Cholesky factorization. More details are provided in the appendix. For experiments on the UCI datasets, we minimize the hinge loss with libSVM to appropriately compare with prior work \cite{arora2020harnessing, fernandez2014we}.

\subsection{MNIST}

As a ``unit test," we evaluate the performance of the compositional kernels in comparison to several baseline methods, 
including the Gaussian kernel, on the MNIST dataset of handwritten digits \cite{lecunmnist}.
Results are presented in Table \ref{table:MNIST}. We observe that all convolutional methods show nearly identical performance, outperforming the three non-convolutional methods (NTK, arccosine kernel, and Gaussian kernel).

\begin{table}[h]
\centering
\caption{Classification performance on MNIST. All methods with convolutional structure have essentially the same performance.} 
\begin{tabular}{l|l}
\hline
Method                                  & \multicolumn{1}{|p{2cm}}{\centering MNIST \\ Accuracy } \\ \hline
NTK                                    & 98.6 \\
ArcCosine Kernel                       & 98.8 \\
Gaussian Kernel                        & 98.8  \\
Gabor Filters + Gaussian Kernel         & 99.4  \\
LeNet-5 \cite{lecun1998gradient}        & 99.2     \\
CKN  \cite{mairal2014convolutional}    & 99.6       \\ \hline \hline
Myrtle5 Kernel                          & 99.5    \\ \hline
Myrtle5 CNN                             & 99.5      
\end{tabular}
\label{table:MNIST}
\end{table}

\subsection{{\cifarten}}
\label{ss:cifarten}

Table \ref{table:cifar10} compares the performance of neural networks with various depths and their corresponding compositional kernels on both the 10,000 test images from {\cifarten} and the additional 2,000 ``harder" test images from {\cifarten}.1\footnote{As this dataset was only recently released, some works do not report accuracy on this dataset.} \cite{krizhevsky2009learning, recht2019imagenet}. 
We include the performance of the Gaussian kernel and a standard ResNet32 as baselines. We train all the Myrtle CNNs on {\cifarten} using SGD and the mean squared error (MSE) loss with multi-step learning rate decay. The exact hyperparameters are provided in the appendix.

We observe that a simple neural network architecture built exclusively from $3 \times 3$ convolutions, $2 \times 2$ average pooling layers, and ReLU nonlinearities, and trained with only flip augmentation, achieves $93\%$ accuracy on {\cifarten}. The corresponding fixed compositional kernel achieves $90\%$ accuracy on the same dataset, outperforming all previous kernel methods. We note the previous best-performing kernel method from \citet{li2019enhanced} heavily relies on a data dependent feature extraction before data is passed into the kernel function \cite{coates2012learning}.
When additional sources of augmentation are used, such as cutout and random crops, the accuracy of the neural network increases to $96\%$.  
Unfortunately due to the quadratic dependence on dataset size, it is currently intractable to augment the compositional kernel to the same extent. For all kernel results\footnote{with the exception of the experiment performed without ZCA processing}  on {\cifarten}, we gained a performance improvement of roughly $0.5\%$ using two techniques: Leave-One-Out tilting and ZCA augmentation we detail these techniques in appendix \ref{app:loozca}.

\paragraph{Effect of preprocessing.} For all of our primary {\cifarten} experiments, we begin with ZCA pre-processing \cite{goodfellow2013maxout}. Table \ref{table:cifar10} also shows the accuracy of our baseline CNN and its corresponding kernel when we replace ZCA with a simpler preprocessing of mean subtraction and standard deviation normalization. We find a substantial drop in accuracy for the compositional kernel without ZCA preprocessing, compared to a much more modest drop in accuracy for the CNN. This result underscores the importance of proper preprocessing for kernel methods; we leave improvements in this area for future work.

\subsection{{\cifarten}0}
For further evaluation, we compute the compositional kernel with the best performance on {\cifarten}  on {\cifarten}0.  
We report our results in Table \ref{table:cifar100}. 
We find the compositional kernel to be modestly performant on {\cifarten}0, matching the accuracy 
of a CNN of the same architecture when no augmentation is used.
However we note this might be due to training instability as the network performed more favorably after
flip augmentation was used. Accuracy further increased when batch normalization was added, lending credence to the training instability
hypothesis.
We also note  cross entropy loss was used to achieve the accuracies in Table \ref{table:cifar100}, as we had difficulty optimizing MSE loss on this dataset.  We leave further investigations on the intricacies of achieving high accuracy on {\cifarten}0 for future work.

\begin{table}[h!]
\centering
\caption{\label{table:cifar100} Accuracy on {\cifarten}0. All CNNs were trained with cross entropy loss.}
\begin{tabular}{l|l}
\hline
Method                          & \multicolumn{1}{|p{2cm}}{\centering {\cifarten}0 \\ Accuracy } \\ \hline
Myrtle10-Gaussian Kernel & 65.3\\
Myrtle10-Gaussian Kernel + Flips & 68.2 \\
Myrtle10 CNN & 64.7 \\
Myrtle10 CNN + Flips & 71.4 \\
Myrtle10 CNN + BatchNorm & 70.3 \\
Myrtle10 CNN + Flips + BatchNorm & 74.7 \\
\end{tabular}
\end{table}

\begin{table*}[h]
\centering
\caption{Classification performance on {\cifarten}.}
\begin{tabular}{l|l|l}
\hline
Method                                  & \multicolumn{1}{|p{2cm}|}{\centering {\cifarten} \\ Accuracy }& \multicolumn{1}{|p{2cm}}{\centering {\cifarten}.1 \\ Accuracy } \\ \hline 
Gaussian Kernel  & 57.4 & -                  \\
CNTK + Flips \cite{li2019enhanced} & 81.4             & -                  \\
CNN-GP + Flips \cite{li2019enhanced} & 82.2             & -                  \\
CKN  \cite{mairal2014convolutional}    & 82.2             & -                  \\  
Coates-NG  + Flips \cite{recht2019imagenet} & 85.6              & 73.1                  \\  
Coates-NG + CNN-GP + Flips \cite{li2019enhanced} & 88.9              & -                  \\  
ResNet32                               & 92.5              & 84.4               \\  \hline \hline
Myrtle5 Kernel + No ZCA               & 77.7 & 62.2 \\
Myrtle5 Kernel                          & 85.8             & 71.6               \\
Myrtle7 Kernel                          & 86.6             & 73.1               \\
Myrtle10 Kernel                         & 87.5             & 74.5               \\
Myrtle10-Gaussian Kernel                     & 88.2             & 75.1               \\
Myrtle10-Gaussian Kernel + Flips  & 89.8             & 78.3              \\ \hline
Myrtle5 CNN + No ZCA                  & 87.8 & 75.8 \\
Myrtle5 CNN                             & 89.8             & 79.0                \\
Myrtle7 CNN                             & 90.2             & 79.7                 \\
Myrtle10 CNN                            & 91.2             & 79.9                 \\
Myrtle10 CNN + Flips        & 93.4             & 84.8                 \\    
Myrtle10 CNN + Flips + CutOut + Crops         & 96.0             & 89.8                 \\    
\end{tabular}
\label{table:cifar10}
\end{table*}

\subsection{Subsampled {\cifarten}}
\begin{figure}[h]
\centering
   \includegraphics[width=0.48\textwidth]{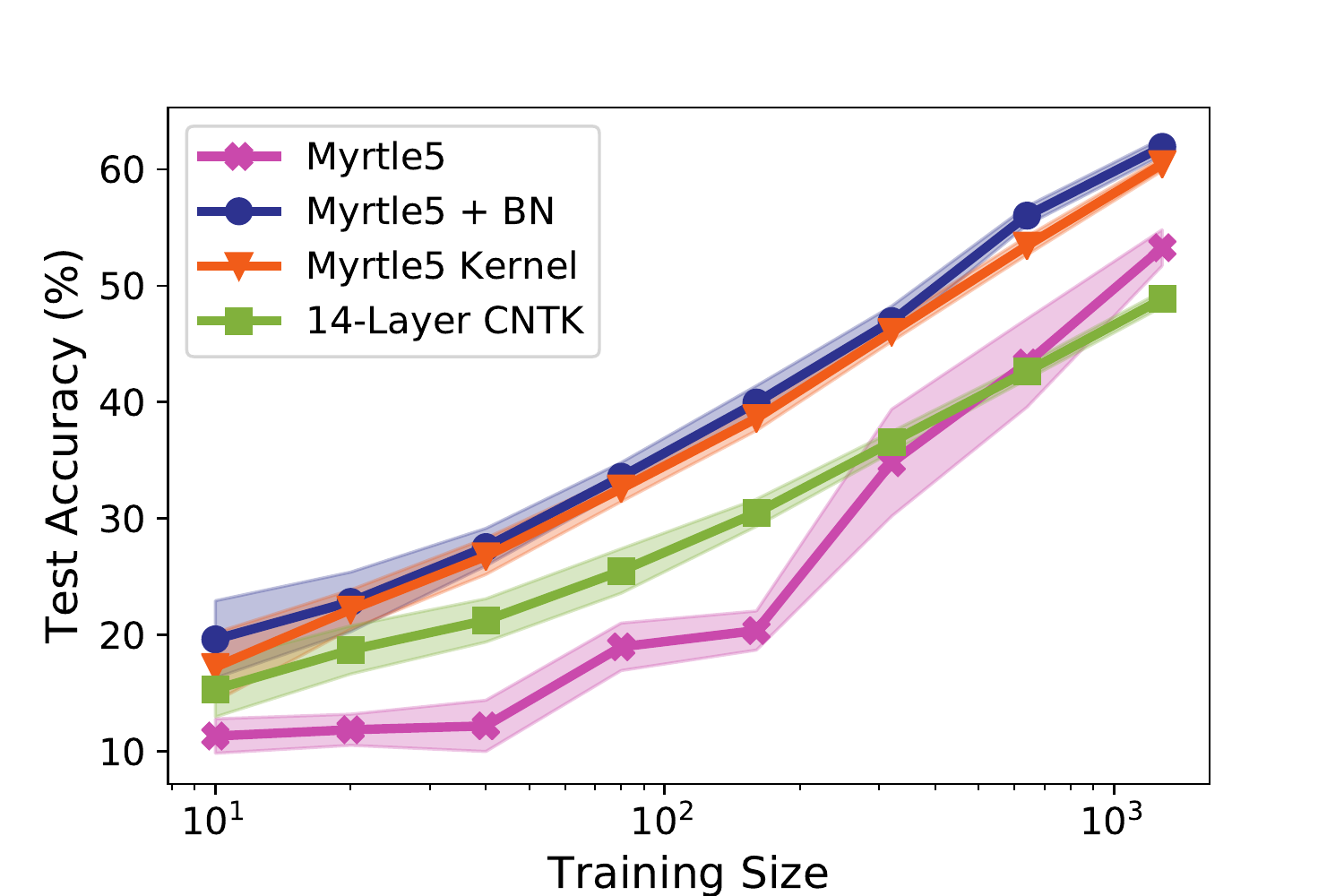}
   \caption{Accuracy results on random subsets of {\cifarten}, with standard deviations over 20 trials. The 14-layer CNTK results are from \citet{arora2020harnessing}.}
   \label{fig:small_cifar10}
\end{figure}

In this section, we present comparison results in the small dataset regime using subsamples of {\cifarten}, 
as investigated in \citet{arora2020harnessing}. Results are shown in Figure \ref{fig:small_cifar10}. Subsampled datasets are class balanced, and standard deviations are computed over 20 random subsamples, as in \citet{arora2020harnessing}. More details are provided in the appendix.

\paragraph{Results.}
We demonstrate that in the small dataset regime explored in \citet{arora2020harnessing}, 
our convolutional kernels significantly outperform the NTK on subsampled training sets of {\cifarten}. 
We find a network with the same architecture as our kernel severely underperforms both the compositional kernel
and NTK in the low data regime. As with \cifarten0{} we suspect this is a training issue as once we
add batch normalization the network outperforms both our kernel and the NTK from \citet{arora2020harnessing}.

\subsection{UCI datasets}

\begin{table*}[h]
\centering
\caption{Results on 90 UCI datasets for the NTK and Gaussian kernel (both tuned over 4 eval folds).}
\begin{tabular}{llllll}  
\toprule
\cmidrule(r){1-2}
Classifier  & Friedman & Average & P90  & P95 & PMA  \\
 & Rank & Accuracy (\%) & (\%) & (\%) & (\%) \\
\midrule
SVM NTK     & 14.3 & 83.2 $\pm$ 13.5 & 96.7 & 83.3 & 97.3 $\pm$ 3.8  \\
SVM Gaussian kernel  & 11.6 & 83.4 $\pm$ 13.4 & 95.6 & 83.3 & 97.5 $\pm$ 3.7  \\
\bottomrule
\end{tabular}
\label{table:UCI_main}
\end{table*}

In this section, we present comparison results between the Gaussian kernel and NTK evaluated on 90 UCI datasets, following the setup used  in \citet{arora2020harnessing}. \citet{arora2020harnessing}  identifies that the NTK outperforms a variety of classifiers, including the Gaussian kernel, random forests (RF), and polynomial kernels, evaluated in \citet{fernandez2014we} on 90 UCI datasets.

\begin{figure}[h]
\centering
   \includegraphics[width=0.48\textwidth]{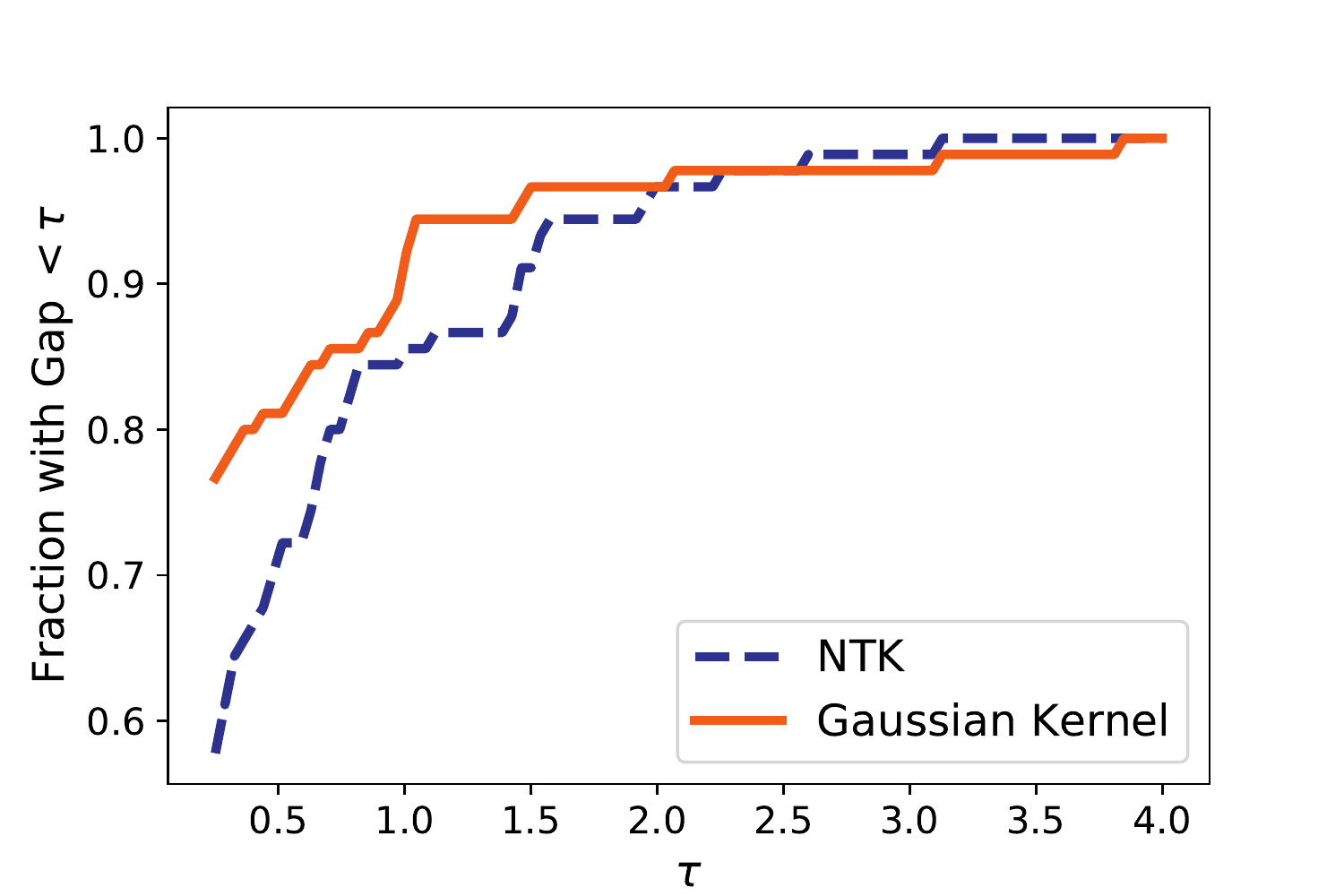}
   \caption{Performance profiles for NTK and tuned Gaussian kernel on 90 UCI datasets.} \label{fig:perf_prof}
\end{figure}

\paragraph{Results.}
For appropriate comparison, we use the same set of 90 ``small" UCI datasets (containing no more than 5000 data points) as in \citet{arora2020harnessing}  for the evaluations. 
For the tuning and evaluation procedure we make one crucial modification to the evaluation procedure posed in \citet{arora2020harnessing} and \citet{fernandez2014we}.
We compute the optimal hyperparameters for each dataset (for both NTK and Gaussian kernel) by averaging performance over \emph{four} cross-validation folds, while both \citet{arora2020harnessing} and \citet{fernandez2014we}
choose optimal hyper parameters on a \emph{single} cross validation fold. 
Using a single cross validation fold can lead to high variance in final performance, especially when evaluation is done purely on small datasets.
A single fold was used in the original experimental setup of \citet{fernandez2014we} for purely computational reasons, and the authors point out the issue of high variance hyperparameter optimization.
Table \ref{table:UCI_main} reports the average cross-validation accuracy over the 90 datasets for the NTK and Gaussian kernel.  Compared to results in \citet{arora2020harnessing}, the modified evaluation protocol increases the performance of both methods, and the gap between the NTK and Gaussian kernel disappears.

We compute the same metrics used in \citet{arora2020harnessing}: Friedman rank, P90, P95 and PMA, where a better classifier is expected to have lower Friedman rank and higher P90, P95, and PMA.  
The average accuracy is reported together with its standard deviation. Friedman rank denotes the ranking metric introduced to compare classifiers across multiple datasets in \citet{demvsar2006statistical}, and reports the average ranking of a given classifier compared to all other classifiers.
P90/P95 denotes the percentage of datasets on which the classifier achieves at least $90\%/95\%$ of the maximum accuracy across all classifiers for this dataset. PMA denotes the average percentage of the maximum accuracy across all classifiers for each dataset. 

On all metrics reported by \citet{arora2020harnessing}, the Gaussian kernel has comparable or better performance relative to the NTK. Figure \ref{fig:perf_prof} shows a performance profile to visually compare the two classifiers~\cite{Dolan02}. For a given $\tau$, the $y$ axis denotes the fraction of instances where a classifier either has the highest accuracy or has accuracy within $\tau$ of the best accuracy. The performance profile reveals that the Gaussian kernel and NTK perform quite comparably on the 90 UCI datasets.


%% file: conclusion.tex
The compositional kernels proposed in this manuscript significantly advance 
the state of the art of kernel methods applied to pattern recognition tasks.
However, these kernels still have significant limitations that must be addressed before they
can be applied in practice.

\paragraph{Computational cost.}
The compositional kernels we study compare all pairs of input pixels for two images with $D$ pixels each,
so the cost of evaluating the kernel function on two data points is $\tilde{O}(D^{2})$.   In addition, $O(N^{2})$ kernel evaluations must be computed to construct the full kernel matrix, creating a total complexity of $\tilde{O}(N^{2}D^{2})$. Even with heavily optimized GPU code, this requires significant computation time. We therefore limited our scope to image datasets with a small pixel count and modest number of examples: {\cifarten}/{\cifarten}0 consist of $60,000$ $32 \times 32 \times 3$ images and MNIST consists of $70,000$ $28 \times 28$ images.
Even with this constraint, the largest compositional kernel matrices we study took approximately $1000$ GPU hours to compute. 
Thus, we believe an imperative direction of future work is reducing the complexity of each kernel evaluation. Random feature methods or other compression schemes could play a significant role here.

Once a kernel matrix is constructed, exact minimization of empirical risk often scales as $O(N^3)$. For datasets with less than 100,000 examples, these calculations can be performed relatively quickly on standard workstations with sufficient RAM. However, even these solves are expensive for larger datasets.
Fortunately, recent work on kernel optimization \cite{ma2018kernel, dai2014scalable, numpywren, Wang19} 
paves a way to scale our approach to larger datasets.

\paragraph{Data augmentation.}
A major advantage  of neural networks is that data augmentation can be added essentially for free. For kernel methods, data augmentation requires treating each augmented example as if it was part of the data set, and hence computation scales superlinearly with the amount of augmentation: if one wants to perform 100 augmentations per example, then the final kernel matrix will be 10,000 times larger, and solving the prediction problem may be one million times slower. Finding new paths to cheaply augment kernels~\cite{Ratner17,Dao19} or to incorporate the symmetries implicit in data augmentation explicitly in kernels should dramatically improve the effectiveness of kernel methods on contemporary datasets. 
One promising avenue is augmentation via kernel ensembling, e.g. by forming many smaller kernels with augmented data and averaging their predictions appropriately.

\paragraph{Architectural modifications.}
We consider a simple set of architectural building blocks (convolution, average pool, and ReLU) in this work, but there exist several commonly used primitives
in deep networks that have no clear analogues for kernel machines (e.g residual connections, max pool, batch normalization, etc.). 
While it is unclear whether these primitives are necessary, the question remains open whether the performance gap between
kernels and neural networks indicates a fundamental limitation of kernel methods
or merely an engineering hurdle that can be overcome (e.g. with improved architectures or by additional subunits).

%% file: acknowledgements.tex
We would like to thank Achal Dave for his insights on accelerating kernel operations for the GPU and Eric Jonas for his guidance on parallelizing our kernel operations with AWS Batch. We would additionally like to thank Rebecca Roelofs, Stephen Tu, Horia Mania, Scott Shenker, and Shivaram Venkataraman for helpful discussions and comments on this work.

This research was generously supported in part by ONR awards N00014-17-1-2191, N00014-17-1-2401, and N00014-18-1-2833, the DARPA Assured Autonomy (FA8750-18-C-0101) and Lagrange (W911NF-16-1-0552) programs, a Siemens Futuremakers Fellowship, an Amazon AWS AI Research Award. SFK was supported by an NSF graduate student fellowship.

%% file: app_background.tex
We proceed with kernel classification as follows. 
Let $C$ be the total number of classes.
Let $\left\{ x_{1} ... x_{N}\right\}$ be $N$ training examples in $d$ dimensions.
Let $\left\{ y_{1} ... y_{N}\right\}$ be $N$ one-hot encoded training labels. We use  $v_{im}$ to denote the $m^{th}$ entry of the vector $v_i$. 
For a choice of kernel function $k(x,y)$, loss function $\mathcal{L}$, and regularization value $\lambda$,
we solve the following optimization problem:

\begin{align}\label{eq:erm}
\underset{\alpha}{\text{minimize}} \frac{1}{N}  \sum_{i=1}^{N}   \mathcal{L}\big(\sum_{j=1}^{N} \mtx{K}_{j\cdot} \alpha, y_{i}\big) + \lambda \operatorname{Tr}(\alpha^T K \alpha)
\end{align}
where $\mtx{K}$ is the matrix of kernel evaluations on the data: $K_{ij} = k(x_i,x_j)$. The prediction for an example $x_{test}$ is:
\begin{align}
\underset{c}{\text{argmax}}  \sum_{i=1}^{N} \alpha_{ic}k(x_{test}, x_{i}) 
\end{align}

If not otherwise specificed, we use the  squared error loss, $\mathcal{L}(\hat{y}, y)  = \|\hat{y}-y\|^2$, for our experiments. In this case, $\alpha$ in \eqref{eq:erm}  is given by
\begin{align*}
	\alpha = (\mtx{K}+\lambda \mtx{I})^{-1} \mtx{Y}
\end{align*}
where $\mtx{Y}$ is the $N \times C$ matrix of all one-hot-encodings of the labels. We allow for the value of $\lambda =0$ in our experiments, and oftentimes this value produces the lowest test error.

%% file: app_loozca.tex
Two additional techniques were used for the Cifar-10 experiments for an additional
$0.5\%$ performance improvement in test accuracy.

\subsection{ZCA Augmentation}

As mentioned in Section \ref{sec:conclusion}, incorporating augmentation directly is difficult for kernel methods.
To capture a small fraction of the benefit of the augmentation in the preprocessing method itself, we first augment the data 20 times using the random augment method proposed in \citet{cubuk2019randaugment}.

We then learn the ZCA preprocessing matrix by computing the eigendecomposition of the augmented training data as described in \cite{goodfellow2013maxout}. 
We then use just the portion of the preprocessed data matrix corresponding to the regular unaugmented training data (or corresponding to the unaugmented training data and its horizontal flips) to compute the kernel matrix.

We find this technique offered a small performance boost of around $0.2\%$ across {\cifarten} and {\cifarten}.

\subsection{Leave-One-Out Tilting}

We additionally found a minor improvement in prediction by averaging the predictions from the true labels and from the labels imputed by leave-one-out prediction. With $\mtx{K}$ and $\mtx{Y}$ defined as they are in section \ref{app:background}, let $\mtx{Q}$ be $(\mtx{K}+\lambda\mtx{I})^{-1}$ and $\mtx{\alpha$} be $(\mtx{K}+\lambda \mtx{I})^{-1}\mtx{Y}$.  $\mtx{\alpha}$ are the coefficients for standard ridge regression.

Let $\mtx{Y_{loo}}$ be a $N \times C$ matrix of leave-one-out predictions. Here, the $i$th row of  $\mtx{Y}_{loo}$ is the output of ridge regression where we predict example $i$ using every element in the entire training set \emph{except} example $i$. For kernel ridge regression, we can actually compute the leave-one-out prediction matrix in closed form:
\[
    Y_{loo_{ic}} = Y_{ic} - \frac{Q_{ii}}{\alpha_{ic}}
\]

Our titled prediction uses an affine combination of the true labels $\mtx{Y}$ and the imputed leave-one-out predictions $\mtx{Y}_{loo}$:
\[
    \mtx{\alpha}_{loo} = (\mtx{K}+\lambda \mtx{I})^{-1}(\mtx{Y} - t\mtx{Y}_{loo})
\]
Where $t$ is chosen to maximize test accuracy on {\cifarten}. We empirically find the optimal value of $t$ to be $0.3$. Though we do not yet have a theoretical justification for this method, we found that this solution never reduced test error, and always performed well on the test set {\cifarten}.1. For our best model, we found this technique offered a modest performance boost of around $0.3\%$ on both {\cifarten} and {\cifarten}.1. We leave an analysis of the efficacy of this technique to future work.

%% file: app_proof.tex
For completeness, we present the proof details for section 3.3 using results from \cite{daniely2016toward}.

We aim to prove the following equality:
\begin{align}
        \begin{split}
        &\mathbb{E}\bigg[ \sum_{c=1}^{D_4} \Psi(\bm{U})[i,j,k,c] \Psi(\bm{U})[\ell,m,n,c]\bigg] =\\ 
        &k_{relu}\Big(c_w\big(k_{0}(\bm{U})\big)\Big)[i,j,k,\ell,m,n]
        \end{split} 
\end{align}
where $\Psi$, $\bm{U}$ and $k_{0}$ are as defined in the main text.

\citet{daniely2016toward} (Section 4.2) presents the concepts of dual activation and kernel:
\begin{align*}
\hat{\sigma}(\rho) = \mathbb{E}_{(X,Y) \sim N_{\rho}} [\sigma(X) \sigma(Y)]
\end{align*}

where we denote by $N_{\rho}$ the multivariate Gaussian distribution with mean 0 and covariance matrix $\begin{pmatrix}
1 & \rho \\
\rho & 1
\end{pmatrix}$.

In our case, we have the \textit{relu} activation $\sigma(\cdot) = \text{max}(x,0)$, whose dual activation function takes the form $\hat{\sigma}(\rho) = \frac{\sqrt{1 - \rho^2} + \rho(\pi - \cos^{-1}(\rho))}{\pi}$ \cite{daniely2016toward}.

Now we show how the above result translates to equation \ref{eq:daniely}. Recall that for a convolutional layer followed by a ReLU layer, we have  tensors $\bm{U}$ with shape $N \times D_1 \times D_2 \times D_3$, and $\bm{W}$  with shape $(2w+1) \times (2w+1) \times D_3 \times D_4$. To ease the notation, denote $\bm{Z}$ as the tensor $\bm{W} * \bm{U}$ with shape $N \times D_1 \times D_2 \times D_4$. We begin with the LHS of equation \ref{eq:daniely}:
\begin{align*}
        \begin{split}
        &\mathbb{E}\bigg[ \sum_{c=1}^{D_4} \Psi(\bm{U})[i,j,k,c] \Psi(\bm{U})[\ell,m,n,c]\bigg] \\
        =& \mathbb{E}\bigg[ \sigma\Big(\sqrt{D_4}(\bm{Z}[i,j,k,1])\Big)\sigma\Big(\sqrt{D_4}(\bm{Z}[l,m,n,1])\Big) \bigg] 
        \end{split} 
\end{align*}
Let $X = \sqrt{D_4}(\bm{Z}[i,j,k,1])$, $Y = \sqrt{D_4}(\bm{Z}[l,m,n,1])$, and choose entries of $W$ to be independent and identically distributed Gaussian random variables with mean 0 and variance $\frac{1}{D_4}$.  With normalization on every patch, we have that $(X,Y)$ follow the multivariate Gaussian distribution with mean 0 and covariance matrix $\begin{pmatrix}
1 & \rho \\
\rho & 1
\end{pmatrix}$, where $$\rho = \sum\limits_{\substack{\delta= -w}}^{w} \sum\limits_{\substack{\Delta= -w}}^{w}\sum\limits_{\substack{d = 1}}^{D_3} U[i,j + \delta, k + \Delta, d]U[l,m + \delta, n + \Delta, d]$$

Following \citet{daniely2016toward}, we have:

\begin{align}
\begin{split}
&\mathbb{E}\bigg[ \sum_{c=1}^{D_4} \Psi(\bm{U})[i,j,k,c] \Psi(\bm{U})[\ell,m,n,c]\bigg] \\
=&\frac{\sqrt{1 - \rho^2} + \rho(\pi - \cos^{-1}(\rho))}{\pi}
\end{split}\label{app-proof-lhs}
\end{align}

Now we show that the RHS of equation \ref{eq:daniely}  is indeed $\frac{\sqrt{1 - \rho^2} + \rho(\pi - \cos^{-1}(\rho))}{\pi}$.

As defined in Subsection \ref{ss:kernels_images}, 
$$k_0(\bm{U})[i,j,k,l,m,n] = \sum_{d = 1}^{D_3}U[i,j,k,d]U[l,m,n,d] $$. 

Let $\bm{C}$ be the tensor $c_w\Big(k_0(\bm{U})\Big)$, 
\begin{align*}
&C[i,j,k,l,m,n]  \\
=& \sum\limits_{\substack{\delta= -w}}^{w} \sum\limits_{\substack{\Delta= -w}}^{w}\sum\limits_{\substack{d = 1}}^{D_3} U[i,j\pm\delta, k \pm \Delta, d]U[l,m\pm\delta, n \pm \Delta, d]
\end{align*}

With normalization for every patch, $\sqrt{C[i,j,k,i,j,k]} = 1$, and
\begin{align}
\begin{split}
&k_{relu}\bigg(c_w\Big(k_0(\bm{U})\Big)\bigg) \\ 
=\quad & \frac{1}{\pi}\bigg(\sqrt{1 - \rho^2}+ \rho(\pi - \cos^{-1}(\rho))\bigg)
\end{split}\label{app-proof-rhs}
\end{align}

Combining (\ref{app-proof-lhs}) and (\ref{app-proof-rhs}) completes the proof.


%% file: app_params.tex
The parameters used to train neural networks for the experiments in this paper are as follows:

For Myrtle5 on MNIST, we used a width of 1,024 filters for all layers and trained for 20 epochs using MSE loss and Adam as the optimizer with a learning rate of 0.001, without weight decay, and without any form of
data preprocessing. For the Myrtle5 Kernel on MNIST we used a regularization value ($\lambda$)  of 1e-4.

For {\cifarten}, all experiments are trained using MSE loss and SGD with Nesterov momentum, setting weight decay to 0.0005, momentum to 0.9, and
minibatch size to 128. All experiments using Myrtle5 used 1,024 filters for all layers and trained for 60 epochs at half-precision with an initial learning rate of 0.1, which is decayed
by 0.1 at 15, 30, and 45 epochs. Myrtle7, Myrtle10 without augmentation, and Myrtle10 with flips used 1,024 filters and are trained for 200 epochs with an initial learning rate
of 0.05, which is decayed by 0.1 at 80, 120, and 160 epochs. Myrtle10 with flips, cutout, and random crops used 2,048 filters and is trained for 400 epochs with an initial learning rate
of 0.1, which is decayed by 0.1 at 80, 160, 240, and 320 epochs. For all {\cifarten} kernel experiments we used a regularization value ($\lambda$) of 0.

For \cifarten0{}, all experiments use a width of 2,048 filters for all layers and are trained for 200 epochs using cross entropy loss and SGD with Nesterov momentum, setting weight decay to 0.0005, momentum to 0.9, and
minibatch size to 128. The learning rate is decayed by 0.2 at 60, 120, and 160 epochs. The initial learning rate is set to 0.1 for both experiments with batch normalization,
0.05 for Myrtle10 CNN with flips, and 0.01 for Myrtle10 CNN without augmentation. For all {\cifarten}0 kernel experiments we used a regularization value ($\lambda$) of 1e-4.

%% file: app_archs.tex
\begin{figure*}[h]
    \begin{subfigure}{0.4\textwidth}
    \centering
    \includegraphics[scale=0.5]{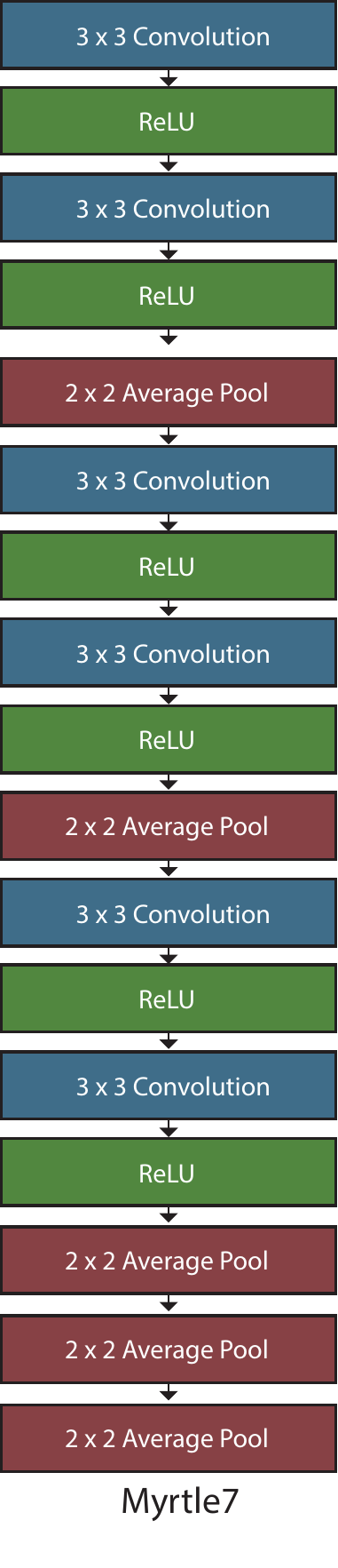}
    \caption{}
    \end{subfigure}
    \begin{subfigure}{0.4\textwidth}
    \centering
    \includegraphics[scale=0.5]{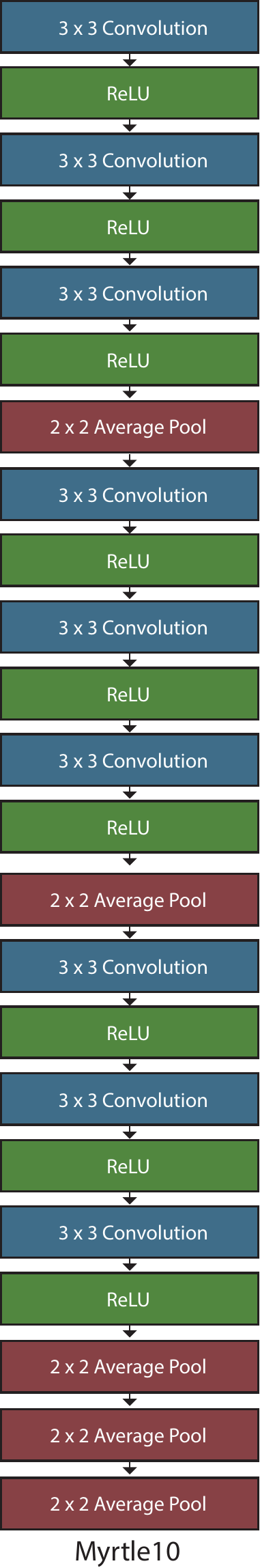}
    \caption{}
    \end{subfigure}
    \caption{\label{fig:archs} a) 7 layer b) 10 layer variants of the Myrtle architectures}
\end{figure*}

In Figure \ref{fig:archs} we illustrate the two ``deeper" Myrtle architectures used. The architectures are similar
to the 5 layer variant illustrated in main text, except with more convolution and nonlinearity layers.

%% file: app_subcifar.tex
Subsets of {\cifarten} were selected uniformly at random without replacement, and each experiment was repeated in 20 independent trials (over which we report standard deviations). This procedure, and the sizes of training sets we consider, match the setup in \cite{arora2020harnessing}. Table \ref{table:small_cifar10} compares the performance (on all 10,000 test examples) of the CNTK from \cite{arora2020harnessing} with that of our Myrtle5 kernel and CNN (with and without batch normalization), our Myrtle10-Gaussian kernel, and a baseline linear classifier. Each linear model used a regularization parameter $\lambda$ tuned on a log scale between $10^{-4}$ and $10^6$. The optimal values for $\lambda$ from top to bottom were: $10^2, 10^{-2}, 10^2, 10^3, 10^3, 10^3, 10^5, 10^4$.  The Myrtle10-Gaussian kernel is our highest-performing unaugmented kernel on the full {\cifarten} dataset; here we confirm that it retains high performance in the small-data regime.

\begin{table*}[h]
\centering
\begin{tabular}{lllllll}  
\toprule
\cmidrule(r){1-2}
Training   &  CNTK  & Myrtle5  & Myrtle5       & Myrtle5  & Myrtle10-G  & Linear \\
Size         &              & CNN      & CNN + BN  & Kernel   & Kernel  & \\
\midrule
10      & $15.33 \pm 2.43$ & $11.29 \pm 1.48$ & $ 19.60 \pm 3.32$ & $17.22 \pm 2.95$ & $19.15  \pm  1.94$ & $12.94 \pm 0.74 $ \\
20      & $18.79 \pm 2.13$ & $11.83 \pm 1.34$ & $ 22.82 \pm 2.56$ & $22.16 \pm 1.69$ & $21.65  \pm  2.97$ & $13.54 \pm 0.69 $ \\
40      & $21.34 \pm 1.91$ & $12.16 \pm 2.20$ & $ 27.53 \pm 1.61$ & $26.74 \pm 1.56$ & $27.20  \pm  1.90$ & $14.66 \pm 0.60 $ \\
80      & $25.48 \pm 1.91$ & $18.96 \pm 2.04$ & $33.58 \pm 1.22$ & $32.56 \pm 1.12$ & $34.22  \pm  1.08$ & $15.54 \pm 0.61 $ \\
160    & $30.48 \pm 1.17$ & $20.36 \pm 1.68$ & $39.96 \pm 1.44$ & $38.61 \pm 1.06$ & $41.89  \pm  1.34$ & $17.15 \pm 0.64 $ \\
320    & $36.57 \pm 0.88$ & $34.79 \pm 4.60$ & $46.96 \pm 1.29$ & $46.03 \pm 0.82$ & $50.06  \pm  1.06$ & $19.18 \pm 0.71 $ \\
640    & $42.63 \pm 0.68$ & $43.36 \pm 3.80$ & $56.03 \pm 0.80$ & $53.45 \pm 0.80$ & $57.60  \pm  0.48$ & $22.30 \pm 0.57 $ \\
1280  & $48.86 \pm 0.68$ & $53.27 \pm 1.55$ & $61.94 \pm 0.74$ & $60.46 \pm 0.58$ & $64.40  \pm  0.48$ & $25.64 \pm 0.61 $ \\
\bottomrule
\end{tabular}
\caption{\label{table:small_cifar10} Accuracy results ($\%$) on random subsets of {\cifarten}, with standard deviations over 20 resamplings. Myrtle10-G denotes the Myrtle10-Gaussian kernel, our best-performing kernel on the full {\cifarten} dataset which retains its high performance in the small-data regime. The shallower Myrtle5 CNN trained with batch normalization has similar performance to the corresponding compositional kernel, both of which outperform the CNTK and the Myrtle5 CNN trained without batch normalization.}
\end{table*}

%% file: app_UCI.tex
In this section, we provide supplementary details for the experiments on the UCI datasets.

Table ~\ref{table:UCI_all} provides the accuracy values (with Clopper-Pearson confidence intervals) for both the NTK and the tuned Gaussian kernel on each of the 90 small UCI datasets. 
For the Gaussian kernel we swept the kernel bandwidth $\gamma$ value from $\nu * 2^{-19}$ to $\nu * 2^{20}$ in log space,
where $\nu$ was chosen heuristically to be the \emph{median} $\ell_2$ distance between data points.
We swept the SVM parameter $C$ from $2^{-19}$ to $2^{20}$ in log space.

For the NTK we tuned the kernel exactly using the protocol from \cite{arora2020harnessing}.

\begin{table*}[h]
\centering
\begin{tabular}{llllll}  
\toprule
\cmidrule(r){1-2}
Classifier  & Friedman & Average & P90  & P95 & PMA  \\
 & Rank & Accuracy (\%) & (\%) & (\%) & (\%) \\
\midrule
SVM NTK (tuned on 4 eval folds)     & 14.3 & 83.2 $\pm$ 13.5 & 96.7 & 83.3 & 97.3 $\pm$ 3.8  \\
SVM Gaussian (tuned on 4 eval folds) & 11.6 & 83.4 $\pm$ 13.4 & 95.6 & 83.3 & 97.5 $\pm$ 3.7  \\
SVM NTK (tuned on 1 eval fold)       & 29.0  & 82.0 $\pm$ 14.0  & 88.9 & 72.2 & 95.6  $\pm$ 5.2  \\
RF (tuned on 1 eval fold)            &  32.4 & 81.6 $\pm$ 13.8 & 85.6 & 65.6 & 95.1 $\pm$ 5.3  \\
SVM Gaussian (tuned on 1 eval fold)  &  35.1  & 81.0 $\pm$ 15.0 & 85.6 & 72.2 & 94.4 $\pm$ 8.2  \\
SVM Polynomial (tuned on 1 eval fold)     & 36.6 & 78.2 $\pm$ 20.2 & 81.6 & 63.2 & 94.3 $\pm$ 6.0  \\
\bottomrule
\label{table:UCI_2}
\end{tabular}
\caption{Results on 90 UCI datasets. Friedman rank reports the average (over datasets) of the accuracy ranking (among classifiers). Average accuracy is reported $\pm$ standard deviation. P90/P95 denotes the percentage of datasets on which a classifier achieves at least $90\%/95\%$ of the maximum accuracy. PMA denotes the average (over datasets) percentage of the maximum accuracy $\pm$ standard deviation. }
\end{table*}

\begin{table*}[h]
\rowcolors{3}{white}{gray!15}
\caption{Results on 90 UCI datasets, with Clopper-Pearson confidence intervals.}
\label{table:UCI_all}
\centering
\begin{tabular}{llllll}  
\toprule
\cmidrule(r){1-2}
Dataset  & NTK Accuracy & Gaussian Kernel Accuracy  \\
\midrule
abalone & 66.04 {\footnotesize \textcolor{gray}{[63.97, 68.08]}} & 66.71 {\footnotesize \textcolor{gray}{[64.65, 68.74]}} \\
acute-inflammation & 100.0 {\footnotesize \textcolor{gray}{[94.04, 100.0]}} & 100.0 {\footnotesize \textcolor{gray}{[94.04, 100.0]}} \\
acute-nephritis & 100.0 {\footnotesize \textcolor{gray}{[94.04, 100.0]}} & 100.0 {\footnotesize \textcolor{gray}{[94.04, 100.0]}} \\
arrhythmia & 72.35 {\footnotesize \textcolor{gray}{[66.02, 78.07]}} & 72.79 {\footnotesize \textcolor{gray}{[66.49, 78.48]}} \\
balance-scale & 99.04 {\footnotesize \textcolor{gray}{[97.22, 99.8]}} & 99.52 {\footnotesize \textcolor{gray}{[97.96, 99.97]}} \\
balloons & 100.0 {\footnotesize \textcolor{gray}{[63.06, 100.0]}} & 100.0 {\footnotesize \textcolor{gray}{[63.06, 100.0]}} \\
bank & 89.93 {\footnotesize \textcolor{gray}{[88.62, 91.14]}} & 90.04 {\footnotesize \textcolor{gray}{[88.74, 91.25]}} \\
blood & 80.48 {\footnotesize \textcolor{gray}{[76.1, 84.38]}} & 79.01 {\footnotesize \textcolor{gray}{[74.53, 83.03]}} \\
breast-cancer & 75.7 {\footnotesize \textcolor{gray}{[67.8, 82.5]}} & 75.35 {\footnotesize \textcolor{gray}{[67.42, 82.19]}} \\
breast-cancer-wisc & 97.86 {\footnotesize \textcolor{gray}{[95.73, 99.1]}} & 97.57 {\footnotesize \textcolor{gray}{[95.36, 98.91]}} \\
breast-cancer-wisc-diag & 97.54 {\footnotesize \textcolor{gray}{[94.99, 99.0]}} & 97.54 {\footnotesize \textcolor{gray}{[94.99, 99.0]}} \\
breast-cancer-wisc-prog & 84.18 {\footnotesize \textcolor{gray}{[75.43, 90.77]}} & 83.16 {\footnotesize \textcolor{gray}{[74.26, 89.97]}} \\
breast-tissue & 75.0 {\footnotesize \textcolor{gray}{[61.05, 85.97]}} & 74.04 {\footnotesize \textcolor{gray}{[60.01, 85.2]}} \\
car & 98.84 {\footnotesize \textcolor{gray}{[97.88, 99.44]}} & 99.31 {\footnotesize \textcolor{gray}{[98.49, 99.74]}} \\
cardiotocography-10clases & 86.53 {\footnotesize \textcolor{gray}{[84.33, 88.53]}} & 85.55 {\footnotesize \textcolor{gray}{[83.29, 87.61]}} \\
cardiotocography-3clases & 93.6 {\footnotesize \textcolor{gray}{[91.95, 94.99]}} & 92.98 {\footnotesize \textcolor{gray}{[91.28, 94.45]}} \\
chess-krvkp & 99.25 {\footnotesize \textcolor{gray}{[98.69, 99.61]}} & 99.22 {\footnotesize \textcolor{gray}{[98.65, 99.59]}} \\
congressional-voting & 63.3 {\footnotesize \textcolor{gray}{[56.53, 69.71]}} & 63.3 {\footnotesize \textcolor{gray}{[56.53, 69.71]}} \\
conn-bench-sonar-mines-rocks & 86.54 {\footnotesize \textcolor{gray}{[78.45, 92.44]}} & 87.98 {\footnotesize \textcolor{gray}{[80.14, 93.54]}} \\
contrac & 54.89 {\footnotesize \textcolor{gray}{[51.21, 58.53]}} & 55.71 {\footnotesize \textcolor{gray}{[52.03, 59.33]}} \\
credit-approval & 87.94 {\footnotesize \textcolor{gray}{[84.02, 91.18]}} & 87.65 {\footnotesize \textcolor{gray}{[83.7, 90.93]}} \\
cylinder-bands & 80.66 {\footnotesize \textcolor{gray}{[75.29, 85.32]}} & 81.05 {\footnotesize \textcolor{gray}{[75.71, 85.67]}} \\
dermatology & 98.08 {\footnotesize \textcolor{gray}{[94.86, 99.53]}} & 98.35 {\footnotesize \textcolor{gray}{[95.26, 99.66]}} \\
echocardiogram & 86.36 {\footnotesize \textcolor{gray}{[75.69, 93.57]}} & 86.36 {\footnotesize \textcolor{gray}{[75.69, 93.57]}} \\
ecoli & 87.5 {\footnotesize \textcolor{gray}{[81.53, 92.09]}} & 88.1 {\footnotesize \textcolor{gray}{[82.21, 92.57]}} \\
energy-y1 & 96.22 {\footnotesize \textcolor{gray}{[93.8, 97.9]}} & 96.48 {\footnotesize \textcolor{gray}{[94.12, 98.09]}} \\
energy-y2 & 89.97 {\footnotesize \textcolor{gray}{[86.52, 92.79]}} & 91.41 {\footnotesize \textcolor{gray}{[88.14, 94.01]}} \\
fertility & 89.0 {\footnotesize \textcolor{gray}{[76.93, 96.08]}} & 89.0 {\footnotesize \textcolor{gray}{[76.93, 96.08]}} \\
flags & 51.56 {\footnotesize \textcolor{gray}{[41.14, 61.89]}} & 54.69 {\footnotesize \textcolor{gray}{[44.2, 64.88]}} \\
glass & 70.75 {\footnotesize \textcolor{gray}{[61.13, 79.19]}} & 72.17 {\footnotesize \textcolor{gray}{[62.62, 80.44]}} \\
haberman-survival & 73.68 {\footnotesize \textcolor{gray}{[65.93, 80.49]}} & 75.66 {\footnotesize \textcolor{gray}{[68.04, 82.25]}} \\
heart-cleveland & 59.87 {\footnotesize \textcolor{gray}{[51.62, 67.73]}} & 60.2 {\footnotesize \textcolor{gray}{[51.95, 68.04]}} \\
heart-hungarian & 87.67 {\footnotesize \textcolor{gray}{[81.22, 92.53]}} & 86.99 {\footnotesize \textcolor{gray}{[80.43, 91.98]}} \\
heart-switzerland & 49.19 {\footnotesize \textcolor{gray}{[36.26, 62.21]}} & 50.0 {\footnotesize \textcolor{gray}{[37.02, 62.98]}} \\
heart-va & 40.0 {\footnotesize \textcolor{gray}{[30.33, 50.28]}} & 41.5 {\footnotesize \textcolor{gray}{[31.73, 51.79]}} \\
hepatitis & 83.97 {\footnotesize \textcolor{gray}{[73.93, 91.31]}} & 86.54 {\footnotesize \textcolor{gray}{[76.92, 93.21]}} \\
ilpd-indian-liver & 74.32 {\footnotesize \textcolor{gray}{[68.9, 79.23]}} & 73.29 {\footnotesize \textcolor{gray}{[67.82, 78.27]}} \\
ionosphere & 95.74 {\footnotesize \textcolor{gray}{[91.61, 98.2]}} & 95.74 {\footnotesize \textcolor{gray}{[91.61, 98.2]}} \\
iris & 98.65 {\footnotesize \textcolor{gray}{[92.7, 99.97]}} & 98.65 {\footnotesize \textcolor{gray}{[92.7, 99.97]}} \\
led-display & 74.6 {\footnotesize \textcolor{gray}{[70.55, 78.36]}} & 75.3 {\footnotesize \textcolor{gray}{[71.28, 79.02]}} \\
lenses & 87.5 {\footnotesize \textcolor{gray}{[56.38, 99.09]}} & 87.5 {\footnotesize \textcolor{gray}{[56.38, 99.09]}} \\
libras & 86.67 {\footnotesize \textcolor{gray}{[80.81, 91.27]}} & 85.83 {\footnotesize \textcolor{gray}{[79.87, 90.57]}} \\
low-res-spect & 93.8 {\footnotesize \textcolor{gray}{[90.19, 96.38]}} & 94.55 {\footnotesize \textcolor{gray}{[91.1, 96.95]}} \\
lung-cancer & 65.62 {\footnotesize \textcolor{gray}{[38.34, 86.94]}} & 62.5 {\footnotesize \textcolor{gray}{[35.43, 84.8]}} \\
lymphography & 89.19 {\footnotesize \textcolor{gray}{[79.8, 95.22]}} & 89.19 {\footnotesize \textcolor{gray}{[79.8, 95.22]}} \\
mammographic & 82.29 {\footnotesize \textcolor{gray}{[78.58, 85.6]}} & 83.75 {\footnotesize \textcolor{gray}{[80.14, 86.94]}} \\
molec-biol-promoter & 91.35 {\footnotesize \textcolor{gray}{[80.2, 97.35]}} & 91.35 {\footnotesize \textcolor{gray}{[80.2, 97.35]}} \\
molec-biol-splice & 87.61 {\footnotesize \textcolor{gray}{[85.89, 89.19]}} & 87.7 {\footnotesize \textcolor{gray}{[85.99, 89.28]}} \\
musk-1 & 90.97 {\footnotesize \textcolor{gray}{[86.58, 94.29]}} & 91.39 {\footnotesize \textcolor{gray}{[87.07, 94.62]}} \\
\bottomrule
\end{tabular}
\end{table*}

\begin{table*}[t]
\rowcolors{3}{white}{gray!15}
\caption{Results on 90 UCI datasets, with Clopper-Pearson confidence intervals: continued.}
\label{table:UCI_all_part2}
\centering
\begin{tabular}{llllll}  
\toprule
\cmidrule(r){1-2}
Dataset  & NTK Accuracy & Gaussian Kernel Accuracy  \\
\midrule
oocytes-merluccius-nucleus-4d & 84.22 {\footnotesize \textcolor{gray}{[80.76, 87.27]}} & 85.78 {\footnotesize \textcolor{gray}{[82.45, 88.7]}} \\
oocytes-merluccius-states-2f & 93.63 {\footnotesize \textcolor{gray}{[91.14, 95.59]}} & 94.12 {\footnotesize \textcolor{gray}{[91.71, 96.0]}} \\
oocytes-trisopterus-nucleus-2f & 86.62 {\footnotesize \textcolor{gray}{[83.15, 89.61]}} & 87.28 {\footnotesize \textcolor{gray}{[83.87, 90.2]}} \\
oocytes-trisopterus-states-5b & 94.41 {\footnotesize \textcolor{gray}{[91.88, 96.33]}} & 95.18 {\footnotesize \textcolor{gray}{[92.79, 96.95]}} \\
ozone & 97.2 {\footnotesize \textcolor{gray}{[96.14, 98.04]}} & 97.36 {\footnotesize \textcolor{gray}{[96.32, 98.17]}} \\
parkinsons & 93.37 {\footnotesize \textcolor{gray}{[86.49, 97.4]}} & 93.37 {\footnotesize \textcolor{gray}{[86.49, 97.4]}} \\
pima & 76.69 {\footnotesize \textcolor{gray}{[72.14, 80.83]}} & 77.34 {\footnotesize \textcolor{gray}{[72.82, 81.44]}} \\
pittsburg-bridges-MATERIAL & 92.31 {\footnotesize \textcolor{gray}{[81.46, 97.86]}} & 94.23 {\footnotesize \textcolor{gray}{[84.05, 98.79]}} \\
pittsburg-bridges-REL-L & 75.0 {\footnotesize \textcolor{gray}{[61.05, 85.97]}} & 75.0 {\footnotesize \textcolor{gray}{[61.05, 85.97]}} \\
pittsburg-bridges-SPAN & 73.91 {\footnotesize \textcolor{gray}{[58.87, 85.73]}} & 73.91 {\footnotesize \textcolor{gray}{[58.87, 85.73]}} \\
pittsburg-bridges-T-OR-D & 89.0 {\footnotesize \textcolor{gray}{[76.93, 96.08]}} & 90.0 {\footnotesize \textcolor{gray}{[78.19, 96.67]}} \\
pittsburg-bridges-TYPE & 72.12 {\footnotesize \textcolor{gray}{[57.95, 83.65]}} & 68.27 {\footnotesize \textcolor{gray}{[53.89, 80.48]}} \\
planning & 71.67 {\footnotesize \textcolor{gray}{[61.19, 80.67]}} & 73.89 {\footnotesize \textcolor{gray}{[63.56, 82.58]}} \\
plant-margin & 84.38 {\footnotesize \textcolor{gray}{[81.67, 86.82]}} & 84.5 {\footnotesize \textcolor{gray}{[81.8, 86.94]}} \\
plant-shape & 74.0 {\footnotesize \textcolor{gray}{[70.81, 77.01]}} & 73.0 {\footnotesize \textcolor{gray}{[69.78, 76.05]}} \\
plant-texture & 85.44 {\footnotesize \textcolor{gray}{[82.8, 87.81]}} & 85.25 {\footnotesize \textcolor{gray}{[82.6, 87.64]}} \\
post-operative & 72.73 {\footnotesize \textcolor{gray}{[57.21, 85.04]}} & 72.73 {\footnotesize \textcolor{gray}{[57.21, 85.04]}} \\
primary-tumor & 53.96 {\footnotesize \textcolor{gray}{[46.02, 61.76]}} & 55.49 {\footnotesize \textcolor{gray}{[47.54, 63.24]}} \\
seeds & 96.15 {\footnotesize \textcolor{gray}{[90.44, 98.94]}} & 96.15 {\footnotesize \textcolor{gray}{[90.44, 98.94]}} \\
semeion & 95.73 {\footnotesize \textcolor{gray}{[94.08, 97.02]}} & 95.85 {\footnotesize \textcolor{gray}{[94.23, 97.13]}} \\
spambase & 94.93 {\footnotesize \textcolor{gray}{[93.96, 95.79]}} & 94.02 {\footnotesize \textcolor{gray}{[92.97, 94.96]}} \\
statlog-australian-credit & 68.31 {\footnotesize \textcolor{gray}{[63.11, 73.2]}} & 68.31 {\footnotesize \textcolor{gray}{[63.11, 73.2]}} \\
statlog-german-credit & 78.4 {\footnotesize \textcolor{gray}{[74.53, 81.93]}} & 78.7 {\footnotesize \textcolor{gray}{[74.85, 82.21]}} \\
statlog-heart & 88.06 {\footnotesize \textcolor{gray}{[81.33, 93.02]}} & 89.18 {\footnotesize \textcolor{gray}{[82.65, 93.88]}} \\
statlog-image & 98.09 {\footnotesize \textcolor{gray}{[97.13, 98.8]}} & 97.66 {\footnotesize \textcolor{gray}{[96.61, 98.45]}} \\
statlog-vehicle & 84.72 {\footnotesize \textcolor{gray}{[80.92, 88.01]}} & 85.31 {\footnotesize \textcolor{gray}{[81.56, 88.55]}} \\
steel-plates & 78.04 {\footnotesize \textcolor{gray}{[75.3, 80.61]}} & 77.53 {\footnotesize \textcolor{gray}{[74.77, 80.12]}} \\
synthetic-control & 99.5 {\footnotesize \textcolor{gray}{[97.88, 99.96]}} & 99.67 {\footnotesize \textcolor{gray}{[98.16, 99.99]}} \\
teaching & 60.53 {\footnotesize \textcolor{gray}{[48.65, 71.56]}} & 61.18 {\footnotesize \textcolor{gray}{[49.31, 72.16]}} \\
tic-tac-toe & 99.79 {\footnotesize \textcolor{gray}{[98.84, 99.99]}} & 100.0 {\footnotesize \textcolor{gray}{[99.23, 100.0]}} \\
titanic & 78.95 {\footnotesize \textcolor{gray}{[76.42, 81.33]}} & 78.95 {\footnotesize \textcolor{gray}{[76.42, 81.33]}} \\
trains & 87.5 {\footnotesize \textcolor{gray}{[28.38, 99.99]}} & 87.5 {\footnotesize \textcolor{gray}{[28.38, 99.99]}} \\
vertebral-column-2clases & 87.34 {\footnotesize \textcolor{gray}{[81.03, 92.15]}} & 87.66 {\footnotesize \textcolor{gray}{[81.41, 92.41]}} \\
vertebral-column-3clases & 84.74 {\footnotesize \textcolor{gray}{[78.07, 90.02]}} & 85.06 {\footnotesize \textcolor{gray}{[78.44, 90.29]}} \\
waveform & 87.16 {\footnotesize \textcolor{gray}{[85.79, 88.45]}} & 86.96 {\footnotesize \textcolor{gray}{[85.58, 88.26]}} \\
waveform-noise & 86.8 {\footnotesize \textcolor{gray}{[85.41, 88.1]}} & 86.8 {\footnotesize \textcolor{gray}{[85.41, 88.1]}} \\
wine & 98.86 {\footnotesize \textcolor{gray}{[93.83, 99.97]}} & 98.3 {\footnotesize \textcolor{gray}{[92.91, 99.88]}} \\
wine-quality-red & 67.38 {\footnotesize \textcolor{gray}{[64.0, 70.62]}} & 65.31 {\footnotesize \textcolor{gray}{[61.9, 68.61]}} \\
wine-quality-white & 67.59 {\footnotesize \textcolor{gray}{[65.69, 69.44]}} & 66.14 {\footnotesize \textcolor{gray}{[64.22, 68.01]}} \\
yeast & 61.05 {\footnotesize \textcolor{gray}{[57.44, 64.58]}} & 61.32 {\footnotesize \textcolor{gray}{[57.71, 64.84]}} \\
zoo & 100.0 {\footnotesize \textcolor{gray}{[92.89, 100.0]}} & 99.0 {\footnotesize \textcolor{gray}{[91.03, 100.0]}} \\
\bottomrule
\end{tabular}
\end{table*}